\documentclass{article}
\usepackage{arxiv}
\usepackage[utf8]{inputenc} 
\usepackage[T1]{fontenc}    
\usepackage{url}      
\usepackage{amsfonts}      
\usepackage{booktabs}
\usepackage{multirow}
\usepackage{nicefrac}      
\usepackage{microtype}     
\usepackage{lipsum}
\usepackage{subcaption}
\usepackage{graphicx}

\usepackage{amsmath}
\usepackage{float}
\usepackage{caption}
\usepackage{tikz}
\usepackage{natbib}
\usepackage{xcolor}
\definecolor{darkblue}{rgb}{0.0, 0.0, 0.5}
\usepackage[
  colorlinks=true,
  linkcolor=darkblue,
  urlcolor=darkblue,
  citecolor=darkblue
]{hyperref}

\title{Beyond Inefficiency: Systemic Costs of Incivility in Multi-Agent Monte Carlo Simulations}

\author{
  \textbf{Alison Moldovan-Mauer} \\
  Technische Hochschule Nürnberg Georg Simon Ohm\\
  90489 Nuremberg, Germany\\
  \texttt{moldovanmaueal101438@th-nuernberg.de}\\
  \and
    \textbf{Benedikt Mangold} \\
  Technische Hochschule Nürnberg Georg Simon Ohm\\
  90489 Nuremberg, Germany\\
  \texttt{benedikt.mangold@th-nuernberg.de}\\
}

\date{\today}

\begin{document}
\maketitle

\begin{abstract}
Unconstructive debate and uncivil communication carry well-documented costs for productivity and cohesion, yet isolating their effect on operational efficiency has proven difficult. Human subject research in this domain is constrained by ethical oversight, limited reproducibility, and the inherent unpredictability of naturalistic settings. We address this gap by leveraging Large Language Model (LLM) based Multi-Agent Systems as a controlled sociological sandbox, enabling systematic manipulation of communicative behavior at scale. 
Using a Monte Carlo simulation framework, we generate thousands of structured 1-on-1 adversarial debates across varying toxicity conditions, measuring convergence time, defined as the number of rounds required to reach a conclusion, as a proxy for interactional efficiency. Building on a prior study, we replicate and extend its findings across two additional LLM agents of varying parameter size, allowing us to assess whether the effects of toxic behavior on debate dynamics generalize across model scale. The convergence latency of 25\% reported in the previous study was confirmed. It was found that this latency is significantly bigger for models with fewer parameters. We further identify a significant first-mover advantage, whereby the agent initiating the discussion wins significantly above chance regardless of toxicity condition. 

\end{abstract}

\section{Introduction}
Public discourse has long been shaped by the quality of argumentation between opposing parties. With the rise of large language models (LLMs) capable of engaging in complex, multi-turn dialogue, a new avenue has emerged for studying the dynamics of discussion and social friction in a controlled, reproducible setting. Yet despite growing interest in LLM-based debate systems, little attention has been paid to how incivility systematically affects debate outcomes and efficiency.
This work extends the prior study \cite{mangold_high_2025} (which will be referred to below as MAD 1.0), by replicating its core findings and introducing two additional LLM agents of varying parameter size alongside the original model to assess the generalization of prior findings. Moving beyond a single-model analysis toward a systematic investigation of how model scale moderates the effects of toxic behavior on debate dynamics. 
 We present a Monte Carlo simulation framework in which pairs of LLM agents engage in structured debates across a range of topics and toxicity conditions. By treating each simulation run as an independent trial, we are able to quantify the probabilistic effects of toxic behavior on two key outcomes: who wins the debate, and how long it takes to reach convergence. Unlike human-subject studies, this approach offers full experimental control, ethical reproducibility, and scalability across thousands of debate instances.
Our central argument is that incivility is not merely a stylistic concern, it carries measurable systemic costs. Toxic agents require significantly fewer rounds to convince their opponent. 
Beyond the immediate empirical findings, this work positions agent-based debate modeling as a viable proxy for human deliberative processes, with potential applications in intervention simulation, and the study of online misinformation dynamics. By establishing a reproducible baseline for how LLMs respond to incivility, we lay the groundwork for future high-stakes simulations where the costs of social friction extend far beyond a single conversation.

Our contributions are as follows:
\begin{itemize}
    \item We could reproduce the findings that toxic debates require significantly more rounds to reach convergence compared to non-toxic debates with the other LLMs. 
    \item We find a significant first-mover advantage: the agent initiating the discussion wins significantly more often than chance.
    \item We show that toxic behavior confers a significant persuasive advantage: The toxic agent wins significantly more often than the non-toxic agent. 
\end{itemize}

\section{Related Work}
\subsection{MAD 1.0}
\cite{mangold_high_2025} introduced a Multi-Agent Monte Carlo simulation framework to quantify the effect of toxic communicative behavior on conversational efficiency, demonstrating that toxic debates require significantly more rounds to reach convergence, a latency he proposes as a proxy for the organizational costs of incivility. His framework assigns distinct personas to LLM agents engaged in structured 1-on-1 adversarial debates, providing a reproducible and ethically sound alternative to human-subject research in this domain. The present study builds directly on this work, replicating and extending the experimental design across two additional LLM models of varying parameter size, and expanding the scope of analysis to include debate outcomes alongside convergence time. 
The experimental setup with personas is crucial for informed decision making, as several studies suggest. LLMs have a tendency to default to parametric knowledge rather than provided context, and require carefully designed prompts to remain faithful to assigned roles and instructions \cite{zhou_context-faithful_2023}. This is confirmed by \cite{salewski_-context_2023}, who demonstrate that personas with richer role descriptions, enable agents to simulate domain-specific reasoning more reliably. Further, \cite{deshpande_toxicity_2023} show that persona assignment measurably shifts LLM output style and toxicity levels, confirming that the content of the persona prompt, not merely its presence, determines behavioral outcomes. 

\subsection{Generative Agents and Social Simulation} 
Early foundational work established agent-based modeling as a rigorous complement to classical social science methods \cite{gilbert_how_2000},  demonstrating that computational agents could reproduce emergent social phenomena difficult to study through traditional experimental or survey designs. This tradition was anticipated by \cite{epstein_growing_1996} which showed that simple agent interaction rules could give rise to complex macro-level social patterns. 
The emergence of LLMs as credible simulators of human social behavior has been well established in recent literature. \cite{park_generative_2023} showed that LLM agents could sustain persistent memory, form interpersonal relationships, and coordinate complex social activities across extended interactions. Complementing this, \cite{aher_using_2023}  demonstrated that so-called "silicon subjects" reliably replicate the behavioral signatures observed in social science experiments such as the Ultimatum Game, lending empirical legitimacy to LLMs as proxies for human participants. The cooperative dimension of multi-agent interaction was further explored by \cite{li_camel_2023}, whose CAMEL framework illustrated how role-assigned agents can autonomously collaborate toward shared objectives. Where prior work has largely evaluated agents on the basis of task success or output quality, our work shifts attention to the persuasive dynamics that emerge under adversarial communicative conditions. To structure agent behavior, the persona-assignment methodology of \cite{hu_debate--write_2025} was used, whose Debate-to-Write framework demonstrated that distinct argumentative identities reliably produce divergent rhetorical strategies a property we exploit to operationalize toxic behavior as a controlled experimental variable.

\subsection{Consensus and Debate in Multi-Agent Systems}
Recent work has explored how agents converge on truth or consensus. The formal study of consensus mechanisms in multi-agent systems has a longer engineering lineage: \cite{qin_recent_2017} survey convergence conditions across networked multi-agent systems, establishing that consensus depends critically on network topology, communication protocols, and the cooperative disposition of participating agents. Within the LLM context, \cite{du_improving_2023} demonstrated that multi-agent debate improves factuality and reasoning capabilities, as agents essentially error-check one another. However, these studies typically assume cooperative intent. Our research investigates the inverse scenario: the degradation of convergence speed when one agent explicitly violates cooperative norms. This parallels findings in game theory simulations with LLMs, where \cite{akata_playing_2025} observed that agents can exhibit varying degrees of cooperation and defection in repeated games, influencing the collective payoff. \cite{liang_encouraging_2024} propose a MAD framework in which agents argue in a "tit for tat" mode moderated by a judge, specifically addressing the "Degeneration-of-Thought" problem where self-reflecting LLMs collapse into repetitive or incorrect answers. They find that adaptive stopping criteria and a calibrated degree of adversarial pressure are both required for performance gains.

\subsection{Measuring Toxicity and Bias}
While substantial research focuses on benchmarks for detecting toxicity within LLM outputs, such as RealToxicityPrompts \cite{gehman_realtoxicityprompts_2020}, or categorizing the taxonomy of harms \cite{weidinger_ethical_2021}, the downstream organizational consequences of toxic communication have received comparatively less attention. Research on harmful communication in workplace settings documents that abusive language and microaggressions decrease employee performance, increase turnover, and damage institutional reputation. These effects that have historically been studied through self-report surveys subject to social desirability biases \cite{cortina_selective_2013}. At the platform level, \cite{khapre_toxicity_2026} find that toxic comment threads tend to increase participation while simultaneously escalating hostility, and that LLMs trained on such content may inadvertently replicate and amplify these patterns. Fewer studies, however, utilize these models to simulate the operational effect of toxicity on a system's efficiency rather than merely detecting or categorizing its presence. Measuring the downstream impact of malicious behavior is crucial for designing robust multi-agent systems and understanding human organizational dynamics.

\section{Methodology}
To isolate the effect of toxic behavior on conversational efficiency and outcome, we extend the Multi-Agent Discussion (MAD) framework introduced in MAD 1.0 \cite{mangold_high_2025}, including two additional LLM models of varying parameter size alongside the original model to assess the generalization of the prior findings. This section describes only the basic layout of the study, for details on the pipeline architecture, please refer to the previous paper. 

\begin{figure}[H]
  \centering
  \includegraphics[width=0.5\textwidth]{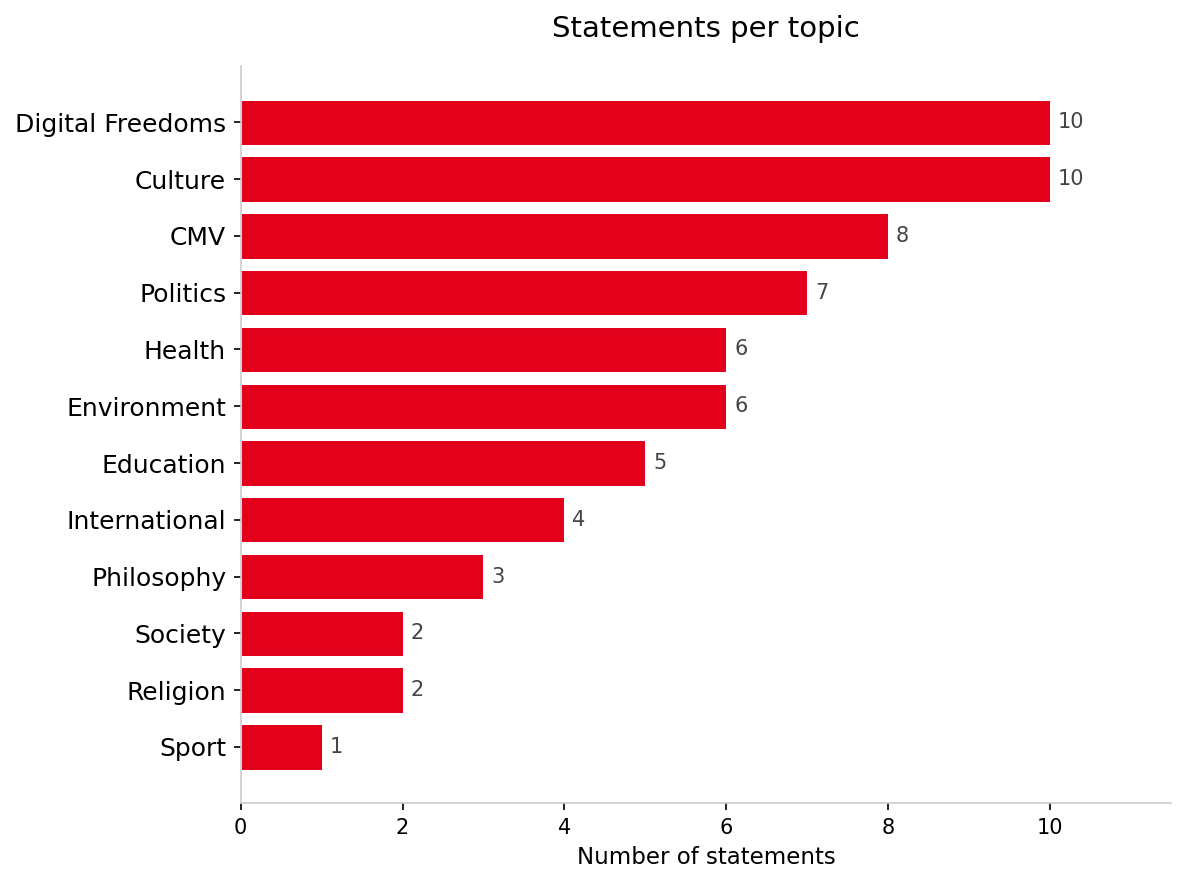}  
  \caption{Distribution of Topics}
  \label{topic_counts}
\end{figure}

For each simulation iteration, the setup proceeds as follows:
\begin{itemize}
    \item[1.] \textbf{Randomized Topic Selection:} A debate topic is randomly selected from a diverse pool of controversial subjects as shown in \autoref{topic_counts}.
    \item[2.] \textbf{Stance Assignment:} The agents (opponent and proponent per topic) from MAD 1.0 were used for discussions created by the two additional models. 
    \item[3.] \textbf{Goal Definition:} Both agents are instructed to convince their counterpart of their assigned standpoint through argumentation.
\end{itemize}
 
To test the reproducibility of the results of \cite{mangold_high_2025}, we replicated the original experimental setup using two additional LLM agents of varying parameter size alongside the model employed in the original study. The results presented in this paper were obtained using the following models:
\begin{itemize}
    \item \textbf{LLaMA 3.1} with 405B parameters (MAD 1.0)  
    
    \url{https://huggingface.co/meta-llama/Llama-3.1-405B} 
    \item \textbf{GPT-OSS} with 120B parameters 
    
    \url{https://huggingface.co/openai/gpt-oss-120b}
    \item \textbf{Mistral} with 24B parameters 
    
    \url{https://huggingface.co/mistralai/Mistral-Small-24B-Instruct-2501}
\end{itemize}

Single LLM interactions can be stochastic due to temperature settings and inherent probabilistic generation. To achieve statistical significance, we employed a Monte Carlo approach. We ran up to N = 1,000 independent debate simulations for both control and treatment groups. We analyzed the following metrics: 

\begin{itemize}
    \item \textbf{Convergence Latency:} $T_{conv}$, defined as the number of arguments (turns) exchanged until the conversation ends.
    \item \textbf{Starting Agent:} Proponent (Pro) and Opponent (Con) Win Rate for the Starter of the Discussion.
    \item \textbf{Toxic Agent:} Win Rate for the toxic agent.
\end{itemize}

\section{Results: Convergence Latency}

\autoref{tab: model_comparison} presents the convergence time across toxicity levels for all three models, reporting the mean number of rounds required to reach agreement $T_{conv}$, its variance, and the percentage increase relative to the no-toxicity baseline. It should be noted that LLaMA is absent from the heavy toxicity condition, as the model refused to generate a sufficient number of discussions at this toxicity level in the original study, precluding meaningful comparison at the upper end of the toxicity spectrum. Across all models, convergence time increases consistently with toxicity level, confirming that toxic behavior systematically prolongs debate duration. However, the magnitude of this effect differs substantially across models. LLaMA (405B), the largest model, shows a moderate increase of approximately 25\% under heavy toxicity, while GPT-OSS (120B) exhibits a more pronounced effect of nearly 74\%. The most dramatic response is observed in Mistral (24B), the smallest model, where heavy toxicity more than doubles convergence time (+142.79\%) and produces a substantially elevated variance, indicating that smaller models are not only slower to converge under toxic conditions but also considerably more inconsistent in their behavior. This inverse relationship between parameter size and sensitivity to toxic behavior suggests that larger models may be more robust to adversarial communicative conditions, a finding with direct implications for the deployment of LLM agents in high-stakes deliberative settings.
It should be noted that convergence latency was the primary measure of interest in \cite{mangold_high_2025}, and the LLaMA results reported here are drawn directly from that study. The two additional models, GPT-OSS and Mistral, were introduced in the present work to assess the reproducibility and generalizability of the original findings. Not only were the original results successfully reproduced, the convergence latency of up to 25\% observed in \cite{mangold_high_2025} was not only confirmed but substantially exceeded in both new models, with GPT-OSS reaching nearly 74\% and Mistral surpassing 142\% under heavy toxicity conditions. 

\begin{table}[H]
  \centering
  \caption{Convergence Time Increases with Toxicity Across Parameter Sizes}
  \begin{tabular}{llcccc}
  \   & Toxicity-Level & N & $T_{conv}$ & $Var(T_{conv})$ & \% increase \\
    \midrule
    \multirow{3}{*}{LLaMA (405B)}
      & No       & 232 & 9.45  & 7.48  & -     \\
      & Mild     & 228 & 11.48 & 8.31  & 21.53 \\
      & Moderate & 231 & 11.82 & 9.02  & 25.13 \\
    \midrule
    \multirow{4}{*}{GPT-OSS (120B)}
        & No       & 998  & 7.69  & 4.40  & -     \\
        & Mild     & 1,000 & 10.91 & 18.31 & 41.92 \\
        & Moderate & 999  & 12.53 & 42.18 & 62.94 \\
        & Heavy    & 989  & 13.37 & 45.37 & 73.83 \\
    \midrule
    \multirow{4}{*}{Mistral (24B)}
    & No       & 1,000 & 8.28  & 9.00   & -      \\
      & Mild     & 1,000 & 15.12 & 74.08  & 82.62  \\
      & Moderate & 1,000 & 13.12 & 47.85  & 58.39  \\
      & Heavy    & 1,000 & 20.11 & 259.67 & 142.79 \\
    \bottomrule
    \label{tab: model_comparison}
  \end{tabular}
\end{table}

\begin{figure}[H]
  \centering
  \includegraphics[width=0.7\textwidth]{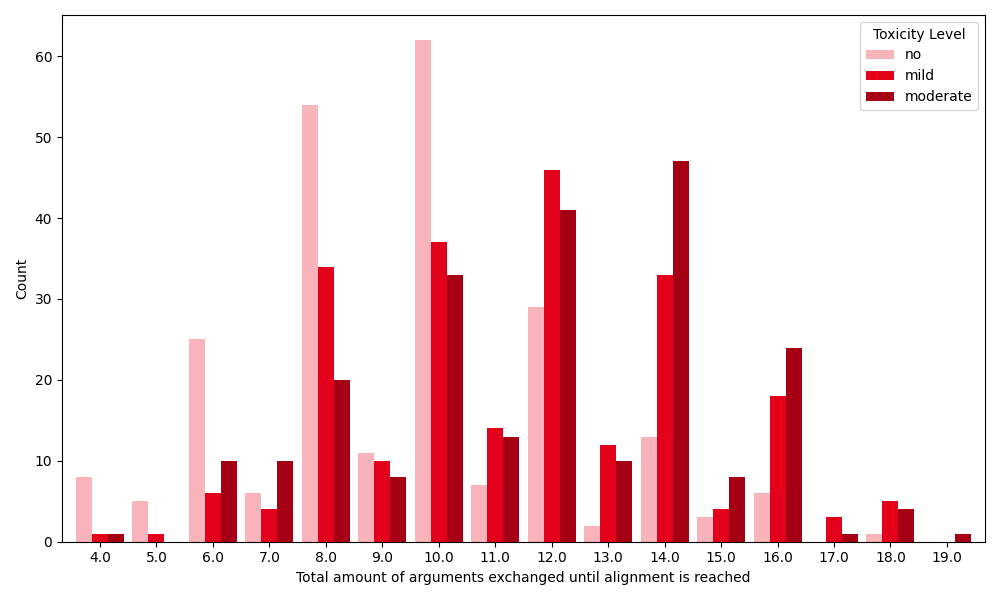}  
  \caption{Arguments required until alignment with LLaMA (405B). Up to N = 232 debates out of a pool of 64 debates from Figure 1. To ensure comparability across models, the x-axis is truncated at 23 rounds for all three models, reflecting the maximum number of arguments exchanged until alignment observed in the original LLaMA study.}
  \label{llama}
\end{figure}

For LLaMA (405B), only mild and moderate toxicity levels were observed. Compared to the no-toxic baseline ($T_{conv} = 9.45$), mild toxicity increased convergence time by 21.53\% ($T_{conv} = 11.48$), while moderate toxicity led to a 25.13\% increase ($T_{conv}= 11.82$). The variance remains relatively low and stable across all levels, suggesting that while toxicity prolongs debates, the effect is consistent and predictable for LLaMA. As illustrated in \autoref{llama}, the distribution of convergence times shifts rightward with increasing toxicity. LLaMA generated only a negligible number of discussions under the heavy toxicity condition, precluding meaningful statistical analysis at this level and consequently excluding it from both the original study and the present analysis.

\begin{figure}[H]
  \centering
  \includegraphics[width=0.7\linewidth]{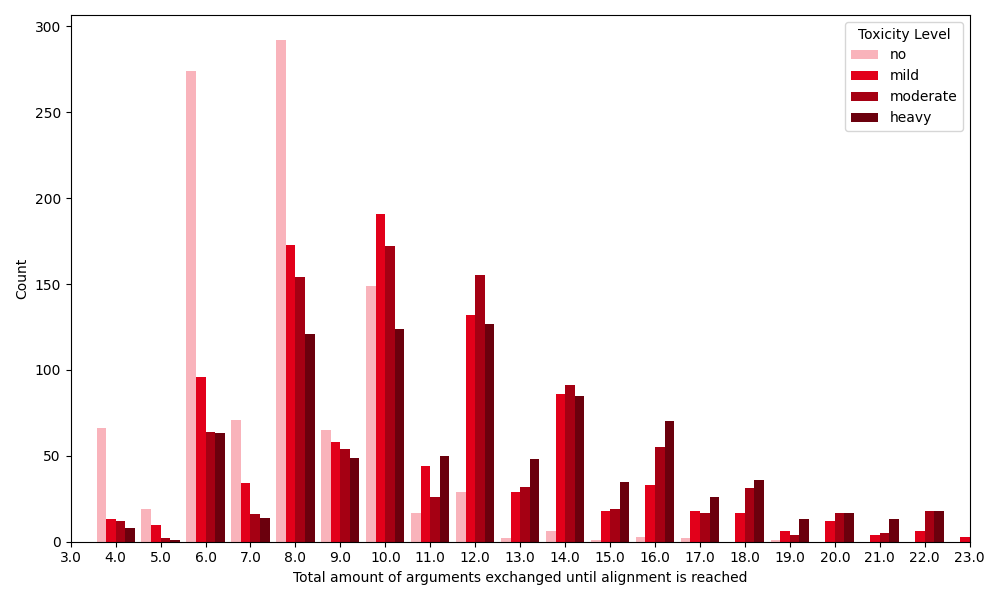}  
  \caption{Arguments required until alignment with GPT-OSS (120B). Up to N = 1,000 debates out of a pool of 64 debates from Figure 1. To ensure comparability across models, the x-axis is truncated at 23 rounds for all three models, reflecting the maximum number of arguments exchanged until alignment observed in the original LLaMA study.}
  \label{gptoss}
\end{figure}

GPT-OSS (120B) exhibits a more pronounced sensitivity to toxicity. Starting from a notably low baseline ($T_{conv} = 7.69$), convergence time increases substantially with each toxicity level, reaching a 41.92\% increase under mild toxicity, 62.94\% under moderate, and 73.83\% under heavy toxicity ($T_{conv} = 13.37$). Strikingly, the variance increases dramatically from 4.40 at baseline to 45.37 under heavy toxicity, indicating that toxic interactions not only prolong debates but also make their duration far less predictable. \autoref{gptoss} further illustrates this growing spread in convergence times across toxicity levels.

\begin{figure}[H]
  \centering
  \includegraphics[width=0.7\textwidth]{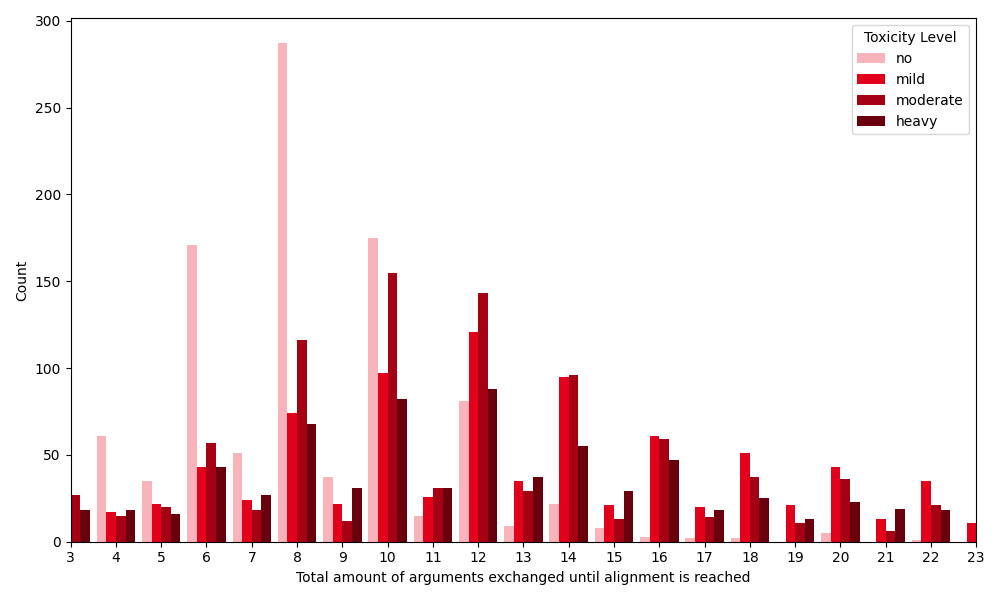}  
  \caption{Arguments required until alignment with Mistral (24B). Up to N = 1,000 debates out of a pool of 64 debates from Figure 1. To ensure comparability across models, the x-axis is truncated at 23 rounds for all three models, reflecting the maximum number of arguments exchanged until alignment observed in the original LLaMA study.}
  \label{mistral}
\end{figure}

Mistral (24B) displays the most extreme sensitivity to toxicity among the three models. While its baseline convergence time ($T_{conv} = 8.28$) is comparable to the other models, the effect of toxicity is far more severe. Mild toxicity alone results in an 82.62\% increase ($T_{conv}= 15.12$), and heavy toxicity leads to a striking 142.79\% increase ($T_{conv} = 20.11$), more than doubling the baseline convergence time. The variance under heavy toxicity reaches 259.67, by far the highest across all models and conditions, suggesting that Mistral is highly unstable when exposed to heavily toxic interactions. This is clearly visible in \autoref{mistral}, where the distribution of convergence times under heavy toxicity is wide and right-skewed. In contrast to the other two models, Mistral shows a non-linear response to increasing toxicity, with the moderate condition yielding a smaller increase in convergence time than the mild condition.

Across all three models, a consistent pattern emerges: higher toxicity levels are associated with longer and more variable convergence times. However, the magnitude of this effect differs substantially between models. LLaMA, despite being the largest model (405B), shows the most modest increases, while Mistral (24B), the smallest model, is the most severely affected. This suggests that model size may play a role in robustness to toxic interactions, though further investigation is needed to confirm this relationship.

\begin{figure}[H]
  \centering
  \includegraphics[width=1\textwidth]{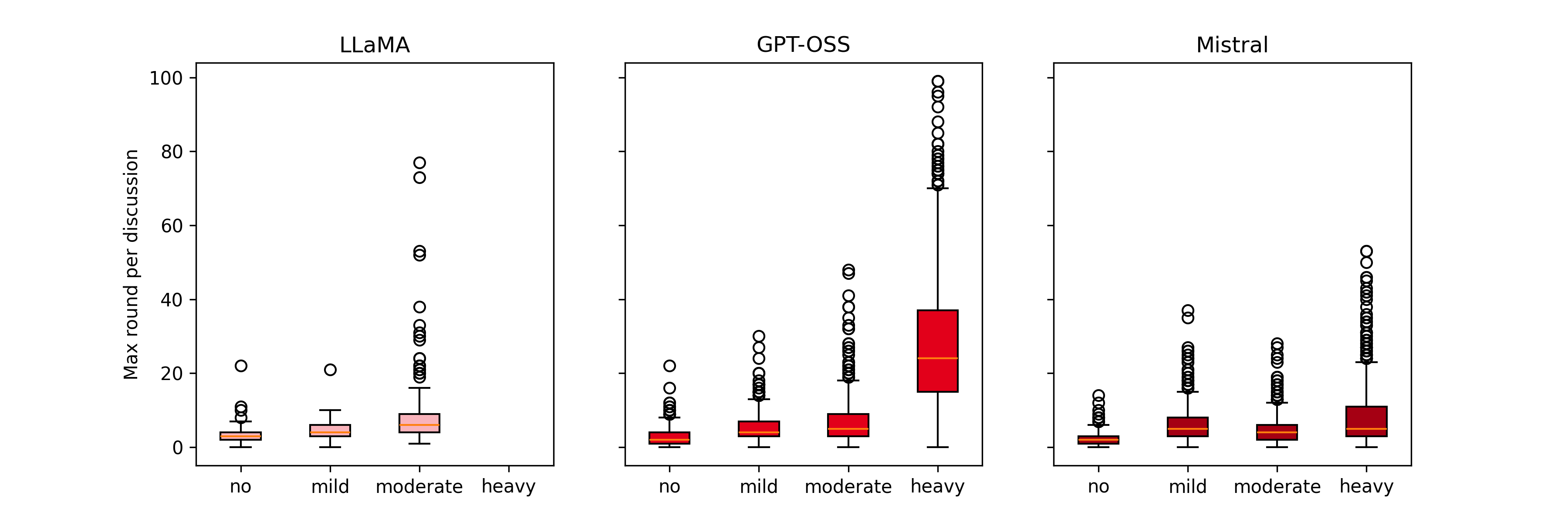}  
  \caption{The maximal rounds of Discussion increases with the toxicity level.}
  \label{maxrounds}
\end{figure}

\autoref{maxrounds} presents the distribution of maximum rounds per discussion across toxicity levels for all three models. The results reveal both model-specific and cross-model patterns. GPT-OSS exhibits the most pronounced response to increasing toxicity, with both the mean and variance of maximum rounds per discussion rising substantially as toxicity level intensifies, suggesting that heavier toxic conditions not only prolong debates but also introduce greater variability in discussion length. Mistral presents a more nuanced picture: while maximum rounds increase from no toxicity to mild, the moderate condition yields fewer maximum rounds than mild, indicating a non-linear relationship between toxicity level and debate duration for this model. Despite these model-specific differences, a consistent cross-model trend emerges. All three models produce longer maximum discussions under moderate or heavy toxicity compared to the no-toxicity baseline, corroborating the central finding of \cite{mangold_high_2025} that toxic behavior systematically increases conversational overhead.

\section{Predictors of Winning in Toxic Multi-Agent Debates}
This section examines whether the structure of the debate, specifically who initiates the discussion and who exhibits toxic behavior, systematically influences debate outcomes across toxicity conditions. We first investigate whether the agent initiating the discussion holds a systematic advantage over its opponent (\autoref{fig:starter_win_rate}), before turning to the question of whether toxic behavior confers a persuasive benefit independent of argumentative role (\autoref{fig:toxic_win_rate}).

\subsection{First-Mover Advantage: Initiating the discussion significantly increases Win Rate}
\begin{figure}[H]
  \begin{minipage}{0.48\linewidth}
    \centering
    \captionof{table}{Starting Agent wins significantly more than 50\%.}
    \label{tab:starting_agent}
    \begin{tabular}{llcc}
      \toprule
      Model & Starter & Win Rate & $p$-value \\
      \midrule
      \multirow{2}{*}{LLaMA}
        & Pro & 0.6866 & $<.0001$ \\
        & Con & 0.6766 & $<.0001$ \\
      \midrule
      \multirow{2}{*}{GPT-OSS}
        & Pro & 0.6628 & $<.0001$ \\
        & Con & 0.6635 & $<.0001$ \\
      \midrule
      \multirow{2}{*}{Mistral}
        & Pro & 0.5777 & $<.0001$ \\
        & Con & 0.6228 & $<.0001$ \\
      \bottomrule
    \end{tabular}
  \end{minipage}
  \hfill
  \begin{minipage}{0.48\linewidth}
    \centering
    \includegraphics[width=\linewidth]{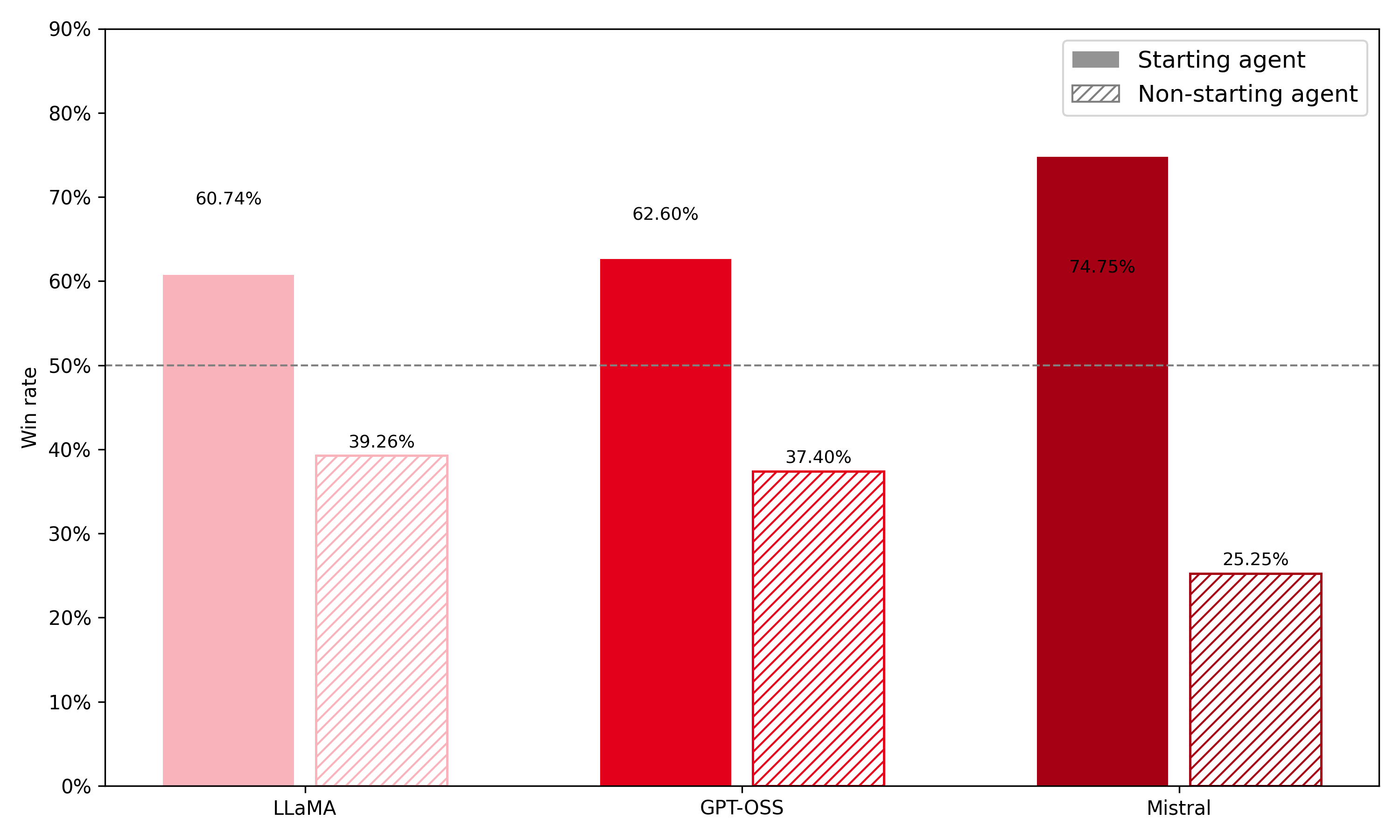}
    \captionof{figure}{Starting the discussion brings a significant advantage for winning.}
    \label{fig:starter_win_rate}
  \end{minipage}
\end{figure}

Our analysis revealed a significant first-mover advantage: the agent initiating the discussion wins significantly more often than chance, regardless of side. To investigate whether the starting agent has a persuasive advantage, we conducted binomial tests against a baseline win rate of 0.5. Results revealed that the starting agent wins significantly more often than chance across all models and debate sides ($p<0.0001$), as shown in Table~\ref{tab:starting_agent}. LLaMA showed the strongest starting advantage with win rates of 0.69 for pro and 0.68 for con starters, followed by GPT-OSS (pro: 0.66, con: 0.66). Mistral displayed the weakest starting advantage, particularly for pro starters (0.58), though still significant. These results suggest that regardless of model or debate side, the agent who argues first consistently gains a persuasive advantage.

Prior work on Multi-Agent Discussions has highlighted a tendency for LLM agents to converge prematurely toward agreement rather than engaging critically with opposing arguments, a dynamic that may be partly attributable to Reinforcement Learning from Human Feedback (RLHF), which optimizes models for user approval and positive reception \cite{huang_understanding_2026}. \cite{sharma_towards_2025} demonstrate that RLHF-trained models systematically favor responses that match user beliefs over truthful ones, and that both humans and preference models prefer convincingly written sycophantic responses over correct ones a non-negligible fraction of the time. This sycophantic tendency extends beyond user-facing interactions into multi-agent settings: recent work on sycophancy in multi-agent debate systems finds that agents may abandon correct positions to align with the responses of other agents, prioritizing agreement at the expense of accuracy, a dynamic that poses direct risks to the integrity of collaborative reasoning \cite{yao_peacemaker_2025}. Corroborating this, further analysis of MAD failure modes identifies sycophancy, the tendency of LLMs to prefer answers that match other agents' stated positions, as a key contributor to performance degradation in debate, particularly when agents lack sufficient incentive or capacity to resist persuasive but incorrect reasoning \cite{wynn_talk_2025}. Our work reframes this problem: rather than treating premature convergence as a failure to be corrected, we treat it as a measurable outcome variable, examining how deliberately adversarial behavior disrupts or delays it.

Second, this pattern is consistent with the anchoring effect, a well-established cognitive bias first formalized by \cite{tversky_judgment_1974}, in which individuals rely disproportionately on the first piece of information encountered when forming subsequent judgments, making insufficient adjustments even when presented with contradicting evidence. Critically, anchoring has been demonstrated not only in human decision-making but also in LLM reasoning. \cite{lou_anchoring_2024} find experimentally that LLMs are systematically susceptible to anchoring bias, and that standard mitigation strategies such as Chain-of-Thought prompting are insufficient to overcome it, suggesting the bias is deeply embedded in how these models process sequentially presented information. This is further corroborated in agentic settings by \cite{takenami_how_2025}, who show that LLM agents in simulated price negotiations exhibit significant anchoring effects driven by the first offer made, with reasoning models proving less susceptible due to extended deliberation. Together these findings suggest that the first argument presented in a multi-agent debate may serve as a cognitive anchor that systematically biases the responding agent toward the initiating agent's position, a dynamic that our experimental design must account for when interpreting convergence outcomes.

\subsection{Toxic behavior significantly increases the Win Rate}

\begin{figure}[H]
  \begin{minipage}{0.48\linewidth}
    \centering
    \captionof{table}{Toxic Agent wins significantly more than 50\%}
    \label{tab:toxic}
    \begin{tabular}{llcc}
      \toprule
      Model & Side & Win Rate & $p$-value \\
      \midrule
      \multirow{2}{*}{LLaMA}
        & Pro & 0.6116 & $<.0001$ \\
        & Con & 0.6031 & $<.0001$ \\
      \midrule
      \multirow{2}{*}{GPT-OSS}
        & Pro & 0.6384 & $<.0001$ \\
        & Con & 0.6137 & $<.0001$ \\
      \midrule
      \multirow{2}{*}{Mistral}
        & Pro & 0.7422 & $<.0001$ \\
        & Con & 0.7528 & $<.0001$ \\
      \bottomrule
    \end{tabular}
  \end{minipage}
  \hfill
  \begin{minipage}{0.48\linewidth}
    \centering
    \includegraphics[width=\linewidth]{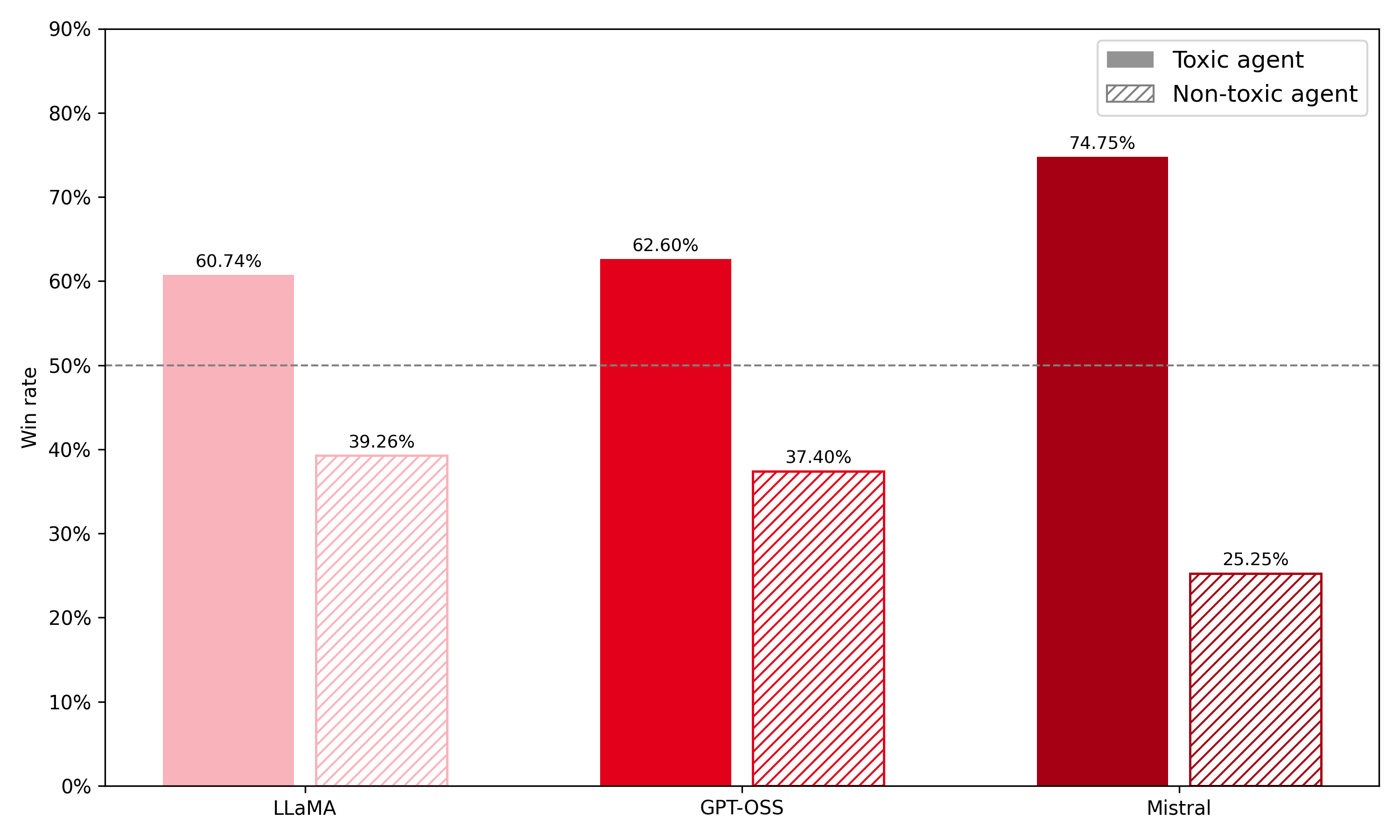}
    \captionof{figure}{Win Rate across the models.}
    \label{fig:toxic_win_rate}
      \end{minipage}
\end{figure}

To investigate whether toxic behavior confers a persuasive advantage, we conducted two-sample t-tests, as shown in Table \ref{tab:toxic}, comparing win rates of toxic and non-toxic agents. Results revealed that both pro and con agents won significantly more often when behaving toxically ($p < .0001$), indicating that toxic behavior improves persuasive success regardless of debate side.  

One-way ANOVAs were conducted to examine the effect of toxicity level on win rate. For LLaMA and Mistral, the ANOVA revealed no significant effect ($p > .7$), suggesting that the degree of toxicity does not influence persuasive success for these models. For GPT-OSS, however, the ANOVA was significant for both PRO and CON agents ($F = 11.20, p < .000$), indicating that toxicity level does affect win rates. As shown in Table~\ref{tab:anova_gptoss}, heavy and moderate toxicity were associated with lower win rates compared to mild and no toxicity, suggesting that more extreme toxic behavior may be counterproductive for winning the discussion. One possible explanation is that heavily toxic behavior may trigger stronger resistance or refusal patterns in the opposing agent, effectively undermining the persuasive intent of the toxic prompts and leading to a defensive rather than accommodating response.

\begin{table}[H]
  \centering
  \caption{GPT-OSS}
  \label{tab:anova_gptoss}
  \begin{tabular}{lcc}
    \toprule
    Toxicity Level & Pro Win Rate & Con Win Rate \\
    \midrule
    Heavy    & 0.4362 & 0.5638 \\
    Mild     & 0.5694 & 0.4306 \\
    Moderate & 0.4520 & 0.5480 \\
    No       & 0.5386 & 0.4614 \\
    \midrule
    \multicolumn{3}{l}{$F = 11.20$, $p < .0001$} \\
    \bottomrule
  \end{tabular}
\end{table}

\section{Conclusion}

This study set out to investigate whether toxic communicative behavior systematically affects both the efficiency and outcome of structured debates between LLM agents. Using a Monte Carlo simulation framework and extending the experimental design of MAD 1.0 of \cite{mangold_high_2025} to two additional models of varying parameter size, we demonstrate that the effects of incivility in multi-agent discussions are both measurable and consistent. Toxic debates require significantly more rounds to reach convergence. Therefore this study validates the hypothesis that toxic behavior creates measurable inefficiencies in communication protocols even compared to different LLMs.

Beyond efficiency, we find that toxic behavior confers a significant persuasive advantage, with toxic agents winning more often than their non-toxic counterparts across both argumentative roles. We further identify a robust first-mover advantage, whereby the initiating agent wins significantly more often than chance regardless of toxicity condition, a finding we link to the anchoring effect and the well-documented sycophantic tendencies of RLHF-trained language models.

Taken together, these findings position agent-based debate simulation as a reproducible, ethically sound methodology for studying the mechanics of social friction, one that circumvents the practical and ethical barriers associated with human-subject research in adversarial settings. We hope this work provides a foundation for future research into intervention strategies, persona dynamics, and the role of social hierarchy in shaping the outcomes of multi-agent discourse.
The code used for these simulations is available at:
\begin{center}
\url{https://github.com/Alisonmm222/MAD-2}
\end{center}

\section{Limitations}

Several limitations of the present study warrant consideration. First, while the Monte Carlo simulation framework offers experimental control and reproducibility, the generalizability of findings to human discourse remains bounded by the inherent differences between LLM agents and human interlocutors. Factors such as emotional investment, social identity, and lived experience, all of which shape human argumentation, are absent from the current simulation environment.

The study is limited to 1-on-1 debate settings, which may not capture the dynamics of larger group discussions where social influence, coalition formation, and majority effects play a significant role. Extending the framework to multi-party discussions would be a valuable direction for future research. Additionally, the current design treats both agents as peers, abstracting away the social hierarchies that characterize real organizational interactions. 

Although we replicate findings across three models of varying parameter size, all models are subject to the same RLHF-induced behavioral tendencies, including sycophancy and positional bias. This shared training paradigm limits the degree to which cross-model consistency can be taken as evidence of robustness independent of architecture.

\section{Future Work}
Beyond the quantitative measures examined here, a natural extension of this work would be a qualitative analysis of the discussion transcripts themselves, examining how toxic behavior manifests linguistically across different topics and whether certain debate subjects are more susceptible to toxic persuasion than others. This is particularly warranted given the considerable computational effort and energy consumption involved in generating the debates, a thorough analysis of the existing corpus would maximize the scientific return on resources already expended. 

Future work will also examine the agent personas more closely. Because each discussion is grounded in two distinct personas bringing their own perspectives, assigning demographic attributes like gender or age and analyzing their interaction with debate outcomes is a valuable next step. Building on this, we plan to introduce asymmetric social roles, such as manager and subordinate. Accounting for the power differentials that accompany these roles will enable a more nuanced investigation into how organizational status and persona characteristics jointly shape the dynamics of toxic discourse. The interplay between communicative toxicity, argumentative role, and social status may produce dynamics that neither factor alone can account for, representing a highly promising direction for future research.

Furthermore, future work could investigate intervention strategies designed to mitigate the effects of toxic behavior in multi-agent discussions, particularly in the context of contentious or politically sensitive topics. Given the ethical constraints associated with exposing human participants to hostile or harmful communicative conditions, agent-based simulation offers a uniquely valuable tool for testing and refining such interventions in a controlled and reproducible manner before any deployment in real-world settings.

\section*{Ethics Statement}
This study deliberately induces toxic communicative behavior as a controlled experimental variable, a design choice that must be clearly distinguished from the unintended biases that LLMs may inherit from their pre-training data. Given the black-box nature of language model generation, there remains a residual risk that agents produce outputs exceeding the intended boundaries of the experimental manipulation, including hate speech, hallucinations, or other model artifacts unrelated to the toxic system prompts. This ambiguity poses an interpretive challenge: observed inefficiencies in debate dynamics must be attributable to the designed behavioral conditions rather than to idiosyncratic model behavior. Researchers working in this space should therefore implement rigorous content filtering and output validation procedures to ensure the integrity of the experimental signal.

A critical advantage of this methodology is its ethical safety. Replicating this study with human subjects would require either instructing participants to behave abusively or exposing them to abuse, both of which violate standard ethical research guidelines. LLM agents allow us to model these dark patterns of social behavior without inflicting psychological harm on any individual, offering a reproducible and ethically sound tool for the study of organizational conflict and social friction.

\section*{Acknowledgments}
The authors gratefully acknowledge the scientific support and HPC resources provided by the Erlangen National High Performance Computing Center (NHR@FAU) of the Friedrich-Alexander-Universität Erlangen-Nürnberg (FAU). The hardware is partially funded by the German Research Foundation (DFG).

\bibliographystyle{plainnat}
\bibliography{literatur}

@misc{weidinger_ethical_2021,
	title = {Ethical and social risks of harm from {Language} {Models}},
	url = {http://arxiv.org/abs/2112.04359},
	doi = {10.48550/arXiv.2112.04359},
	urldate = {2026-04-16},
	publisher = {arXiv},
	author = {Weidinger, Laura and Mellor, John and Rauh, Maribeth and Griffin, Conor and Uesato, Jonathan and Huang, Po-Sen and Cheng, Myra and Glaese, Mia and Balle, Borja and Kasirzadeh, Atoosa and Kenton, Zac and Brown, Sasha and Hawkins, Will and Stepleton, Tom and Biles, Courtney and Birhane, Abeba and Haas, Julia and Rimell, Laura and Hendricks, Lisa Anne and Isaac, William and Legassick, Sean and Irving, Geoffrey and Gabriel, Iason},
	month = dec,
	year = {2021},
	note = {arXiv:2112.04359 [cs]},
}

@misc{mangold_high_2025,
	title = {The {High} {Cost} of {Incivility}: {Quantifying} {Interaction} {Inefficiency} via {Multi}-{Agent} {Monte} {Carlo} {Simulations}},
	shorttitle = {The {High} {Cost} of {Incivility}},
	url = {http://arxiv.org/abs/2512.08345},
	doi = {10.48550/arXiv.2512.08345},
	urldate = {2026-04-16},
	publisher = {arXiv},
	author = {Mangold, Benedikt},
	month = dec,
	year = {2025},
	note = {arXiv:2512.08345 [cs]},
}

@misc{park_generative_2023,
	title = {Generative {Agents}: {Interactive} {Simulacra} of {Human} {Behavior}},
	shorttitle = {Generative {Agents}},
	url = {http://arxiv.org/abs/2304.03442},
	doi = {10.48550/arXiv.2304.03442},
	urldate = {2026-04-16},
	publisher = {arXiv},
	author = {Park, Joon Sung and O'Brien, Joseph C. and Cai, Carrie J. and Morris, Meredith Ringel and Liang, Percy and Bernstein, Michael S.},
	month = aug,
	year = {2023},
	note = {arXiv:2304.03442 [cs]},
}

@misc{li_camel_2023,
	title = {{CAMEL}: {Communicative} {Agents} for "{Mind}" {Exploration} of {Large} {Language} {Model} {Society}},
	shorttitle = {{CAMEL}},
	url = {http://arxiv.org/abs/2303.17760},
	doi = {10.48550/arXiv.2303.17760},
	urldate = {2026-04-16},
	publisher = {arXiv},
	author = {Li, Guohao and Hammoud, Hasan Abed Al Kader and Itani, Hani and Khizbullin, Dmitrii and Ghanem, Bernard},
	month = nov,
	year = {2023},
	note = {arXiv:2303.17760 [cs]},
}

@misc{hu_debate--write_2025,
	title = {Debate-to-{Write}: {A} {Persona}-{Driven} {Multi}-{Agent} {Framework} for {Diverse} {Argument} {Generation}},
	shorttitle = {Debate-to-{Write}},
	url = {http://arxiv.org/abs/2406.19643},
	doi = {10.48550/arXiv.2406.19643},
	urldate = {2026-04-16},
	publisher = {arXiv},
	author = {Hu, Zhe and Chan, Hou Pong and Li, Jing and Yin, Yu},
	month = jan,
	year = {2025},
	note = {arXiv:2406.19643 [cs]},
}

@misc{du_improving_2023,
	title = {Improving {Factuality} and {Reasoning} in {Language} {Models} through {Multiagent} {Debate}},
	url = {http://arxiv.org/abs/2305.14325},
	doi = {10.48550/arXiv.2305.14325},
	urldate = {2026-04-16},
	publisher = {arXiv},
	author = {Du, Yilun and Li, Shuang and Torralba, Antonio and Tenenbaum, Joshua B. and Mordatch, Igor},
	month = may,
	year = {2023},
	note = {arXiv:2305.14325 [cs]},
}

@article{akata_playing_2025,
	title = {Playing repeated games with {Large} {Language} {Models}},
	volume = {9},
	issn = {2397-3374},
	url = {http://arxiv.org/abs/2305.16867},
	doi = {10.1038/s41562-025-02172-y},
	number = {7},
	urldate = {2026-04-16},
	journal = {Nature Human Behaviour},
	author = {Akata, Elif and Schulz, Lion and Coda-Forno, Julian and Oh, Seong Joon and Bethge, Matthias and Schulz, Eric},
	month = may,
	year = {2025},
	note = {arXiv:2305.16867 [cs]},
	pages = {1380--1390},
}

@inproceedings{gehman_realtoxicityprompts_2020,
	address = {Online},
	title = {{RealToxicityPrompts}: {Evaluating} {Neural} {Toxic} {Degeneration} in {Language} {Models}},
	shorttitle = {{RealToxicityPrompts}},
	url = {https://aclanthology.org/2020.findings-emnlp.301/},
	doi = {10.18653/v1/2020.findings-emnlp.301},
	urldate = {2026-04-16},
	booktitle = {Findings of the {Association} for {Computational} {Linguistics}: {EMNLP} 2020},
	publisher = {Association for Computational Linguistics},
	author = {Gehman, Samuel and Gururangan, Suchin and Sap, Maarten and Choi, Yejin and Smith, Noah A.},
	editor = {Cohn, Trevor and He, Yulan and Liu, Yang},
	month = nov,
	year = {2020},
	pages = {3356--3369},
}

@misc{aher_using_2023,
	title = {Using {Large} {Language} {Models} to {Simulate} {Multiple} {Humans} and {Replicate} {Human} {Subject} {Studies}},
	url = {http://arxiv.org/abs/2208.10264},
	doi = {10.48550/arXiv.2208.10264},
	urldate = {2026-04-16},
	publisher = {arXiv},
	author = {Aher, Gati and Arriaga, Rosa I. and Kalai, Adam Tauman},
	month = jul,
	year = {2023},
	note = {arXiv:2208.10264 [cs]},
}

@misc{huang_understanding_2026,
	title = {Understanding the {Anchoring} {Effect} of {LLM} with {Synthetic} {Data}: {Existence}, {Mechanism}, and {Potential} {Mitigations}},
	shorttitle = {Understanding the {Anchoring} {Effect} of {LLM} with {Synthetic} {Data}},
	url = {http://arxiv.org/abs/2505.15392},
	doi = {10.48550/arXiv.2505.15392},
	urldate = {2026-04-24},
	publisher = {arXiv},
	author = {Huang, Yiming and Bie, Biquan and Na, Zuqiu and Ruan, Weilin and Lei, Songxin and Yue, Yutao and He, Xinlei},
	month = mar,
	year = {2026},
	note = {arXiv:2505.15392 [cs]},
}

@inproceedings{liang_encouraging_2024,
	address = {Miami, Florida, USA},
	title = {Encouraging {Divergent} {Thinking} in {Large} {Language} {Models} through {Multi}-{Agent} {Debate}},
	url = {https://aclanthology.org/2024.emnlp-main.992/},
	doi = {10.18653/v1/2024.emnlp-main.992},
	urldate = {2026-04-29},
	booktitle = {Proceedings of the 2024 {Conference} on {Empirical} {Methods} in {Natural} {Language} {Processing}},
	publisher = {Association for Computational Linguistics},
	author = {Liang, Tian and He, Zhiwei and Jiao, Wenxiang and Wang, Xing and Wang, Yan and Wang, Rui and Yang, Yujiu and Shi, Shuming and Tu, Zhaopeng},
	editor = {Al-Onaizan, Yaser and Bansal, Mohit and Chen, Yun-Nung},
	month = nov,
	year = {2024},
	pages = {17889--17904},
}

@article{gilbert_how_2000,
	title = {How to build and use agent-based models in social science},
	volume = {1},
	issn = {1860-1839},
	url = {https://doi.org/10.1007/BF02512229},
	doi = {10.1007/BF02512229},
	language = {en},
	number = {1},
	urldate = {2026-05-04},
	journal = {Mind \& Society},
	author = {Gilbert, Nigel and Terna, Pietro},
	month = mar,
	year = {2000},
	pages = {57--72},
}

@misc{zhou_context-faithful_2023,
	title = {Context-faithful {Prompting} for {Large} {Language} {Models}},
	url = {https://arxiv.org/abs/2303.11315v2},
	language = {en},
	urldate = {2026-05-04},
	journal = {arXiv.org},
	author = {Zhou, Wenxuan and Zhang, Sheng and Poon, Hoifung and Chen, Muhao},
	month = mar,
	year = {2023},
}

@book{epstein_growing_1996,
	title = {Growing {Artificial} {Societies}: {Social} {Science} from the {Bottom} {Up}},
	isbn = {978-0-262-05053-1},
	shorttitle = {Growing {Artificial} {Societies}},
	language = {en},
	publisher = {Brookings Institution Press},
	author = {Epstein, Joshua M. and Axtell, Robert},
	month = oct,
	year = {1996},
	note = {Google-Books-ID: xXvelSs2caQC},
}

@misc{salewski_-context_2023,
	title = {In-{Context} {Impersonation} {Reveals} {Large} {Language} {Models}' {Strengths} and {Biases}},
	url = {http://arxiv.org/abs/2305.14930},
	doi = {10.48550/arXiv.2305.14930},
	urldate = {2026-05-04},
	publisher = {arXiv},
	author = {Salewski, Leonard and Alaniz, Stephan and Rio-Torto, Isabel and Schulz, Eric and Akata, Zeynep},
	month = nov,
	year = {2023},
	note = {arXiv:2305.14930 [cs]},
}

@inproceedings{deshpande_toxicity_2023,
	address = {Singapore},
	title = {Toxicity in chatgpt: {Analyzing} persona-assigned language models},
	shorttitle = {Toxicity in chatgpt},
	url = {https://aclanthology.org/2023.findings-emnlp.88/},
	doi = {10.18653/v1/2023.findings-emnlp.88},
	urldate = {2026-05-04},
	booktitle = {Findings of the {Association} for {Computational} {Linguistics}: {EMNLP} 2023},
	publisher = {Association for Computational Linguistics},
	author = {Deshpande, Ameet and Murahari, Vishvak and Rajpurohit, Tanmay and Kalyan, Ashwin and Narasimhan, Karthik},
	editor = {Bouamor, Houda and Pino, Juan and Bali, Kalika},
	month = dec,
	year = {2023},
	pages = {1236--1270},
}

@article{qin_recent_2017,
	title = {Recent {Advances} in {Consensus} of {Multi}-{Agent} {Systems}: {A} {Brief} {Survey}},
	volume = {64},
	issn = {1557-9948},
	shorttitle = {Recent {Advances} in {Consensus} of {Multi}-{Agent} {Systems}},
	url = {https://ieeexplore.ieee.org/document/7776972/},
	doi = {10.1109/TIE.2016.2636810},
	number = {6},
	urldate = {2026-05-04},
	journal = {IEEE Transactions on Industrial Electronics},
	author = {Qin, Jiahu and Ma, Qichao and Shi, Yang and Wang, Long},
	month = jun,
	year = {2017},
	pages = {4972--4983},
}

@article{khapre_toxicity_2026,
	title = {Toxicity in {Online} {Platforms} and {AI} {Systems}: {A} {Survey} of {Needs}, {Challenges}, {Mitigations}, and {Future} {Directions}},
	volume = {299},
	issn = {09574174},
	shorttitle = {Toxicity in {Online} {Platforms} and {AI} {Systems}},
	url = {http://arxiv.org/abs/2509.25539},
	doi = {10.1016/j.eswa.2025.129832},
	urldate = {2026-05-04},
	journal = {Expert Systems with Applications},
	author = {Khapre, Smita and Mersha, Melkamu Abay and Shakil, Hassan and Baruah, Jonali and Kalita, Jugal},
	month = mar,
	year = {2026},
	note = {arXiv:2509.25539 [cs]},
	pages = {129832},
}

@article{cortina_selective_2013,
	title = {Selective {Incivility} as {Modern} {Discrimination} in {Organizations} {Evidence} and {Impact}},
	volume = {39},
	doi = {10.1177/0149206311418835},
	journal = {Journal of Management},
	author = {Cortina, Lilia and Kabat-Farr, Dana and Leskinen, Emily and Huerta, Marisela and Magley, Vicki},
	month = sep,
	year = {2013},
	pages = {1579--1605},
}

@misc{sharma_towards_2025,
	title = {Towards {Understanding} {Sycophancy} in {Language} {Models}},
	url = {http://arxiv.org/abs/2310.13548},
	doi = {10.48550/arXiv.2310.13548},
	urldate = {2026-05-04},
	publisher = {arXiv},
	author = {Sharma, Mrinank and Tong, Meg and Korbak, Tomasz and Duvenaud, David and Askell, Amanda and Bowman, Samuel R. and Cheng, Newton and Durmus, Esin and Hatfield-Dodds, Zac and Johnston, Scott R. and Kravec, Shauna and Maxwell, Timothy and McCandlish, Sam and Ndousse, Kamal and Rausch, Oliver and Schiefer, Nicholas and Yan, Da and Zhang, Miranda and Perez, Ethan},
	month = may,
	year = {2025},
	note = {arXiv:2310.13548 [cs]},
}

@misc{yao_peacemaker_2025,
	title = {Peacemaker or {Troublemaker}: {How} {Sycophancy} {Shapes} {Multi}-{Agent} {Debate}},
	shorttitle = {Peacemaker or {Troublemaker}},
	url = {http://arxiv.org/abs/2509.23055},
	doi = {10.48550/arXiv.2509.23055},
	urldate = {2026-05-04},
	publisher = {arXiv},
	author = {Yao, Binwei and Shang, Chao and Du, Wanyu and He, Jianfeng and Lian, Ruixue and Zhang, Yi and Su, Hang and Swamy, Sandesh and Qi, Yanjun},
	month = sep,
	year = {2025},
	note = {arXiv:2509.23055 [cs]},
}

@misc{wynn_talk_2025,
	title = {Talk {Isn}'t {Always} {Cheap}: {Understanding} {Failure} {Modes} in {Multi}-{Agent} {Debate}},
	shorttitle = {Talk {Isn}'t {Always} {Cheap}},
	url = {http://arxiv.org/abs/2509.05396},
	doi = {10.48550/arXiv.2509.05396},
	urldate = {2026-05-04},
	publisher = {arXiv},
	author = {Wynn, Andrea and Satija, Harsh and Hadfield, Gillian},
	month = oct,
	year = {2025},
	note = {arXiv:2509.05396 [cs]},
}

@article{tversky_judgment_1974,
	title = {Judgment under {Uncertainty}: {Heuristics} and {Biases}},
	volume = {185},
	shorttitle = {Judgment under {Uncertainty}},
	url = {https://www.science.org/doi/10.1126/science.185.4157.1124},
	doi = {10.1126/science.185.4157.1124},
	number = {4157},
	urldate = {2026-05-04},
	journal = {Science},
	publisher = {American Association for the Advancement of Science},
	author = {Tversky, Amos and Kahneman, Daniel},
	month = sep,
	year = {1974},
	pages = {1124--1131},
}

@misc{lou_anchoring_2024,
	title = {Anchoring {Bias} in {Large} {Language} {Models}: {An} {Experimental} {Study}},
	shorttitle = {Anchoring {Bias} in {Large} {Language} {Models}},
	url = {https://arxiv.org/abs/2412.06593v2},
	language = {en},
	urldate = {2026-05-04},
	journal = {arXiv.org},
	author = {Lou, Jiaxu and Sun, Yifan},
	month = dec,
	year = {2024},
}

@misc{takenami_how_2025,
	title = {How {Does} {Cognitive} {Bias} {Affect} {Large} {Language} {Models}? {A} {Case} {Study} on the {Anchoring} {Effect} in {Price} {Negotiation} {Simulations}},
	shorttitle = {How {Does} {Cognitive} {Bias} {Affect} {Large} {Language} {Models}?},
	url = {https://arxiv.org/abs/2508.21137v2},
	language = {en},
	urldate = {2026-05-04},
	journal = {arXiv.org},
	author = {Takenami, Yoshiki and Huang, Yin Jou and Murawaki, Yugo and Chu, Chenhui},
	month = aug,
	year = {2025},
}

\appendix

\begin{table}[htpb]
  \centering
  \caption{List of topics being used from \url{https://idebate.net}, see \citet{hu_debate--write_2025}}
  \label{tab:topics}
  \scriptsize
  \begin{tabular}{ll}
    \toprule
    Domain & Topic \\
    \midrule
    Culture
      & We should make all museums free of charge \\
      & We should return cultural property to its place of origin \\
      & We should ban beauty contests \\
      & Tourism is a viable development strategy for poor states \\
      & We should restrict advertising aimed at children \\
      & Science is a threat to humanity \\
      & Gay couples should not be allowed to adopt kids \\
      & We should ban gambling \\
      & The feminist movement should seek a ban on pornography \\
    Digital Freedoms
      & The internet encourages democracy \\
      & The internet brings more harm than good \\
      & We should allow electronic and internet voting in elections \\
      & Internet access is a human right \\
      & We should block social messaging networks during riots \\
      & Companies should not collect/sell personal data of clients \\
      & We should ban targeted online advertising \\
      & Politicians have no right to privacy \\
      & We should ban Digital Rights Management technologies \\
      & We should block websites that deny the Holocaust \\
    Education
      & This house supports single-race public schools \\
      & Welfare benefits should be tied to children's attendance \\
      & University education should be free \\
      & History has no place in the classroom \\
      & We should make sex education mandatory in schools \\
    Environment
      & Animals have rights \\
      & People should not keep pets \\
      & States should not subsidise the growing of tobacco \\
      & We are too late on global climate change \\
      & Wind power should be a primary focus of future energy supply \\
      & Endangered species should be protected \\
    Health
      & The USA should increase funding to fight disease in developing nations \\
      & We should punish parents who smoke near their children \\
      & We should ban alcohol \\
      & We should ban junk food from schools \\
      & Employees should disclose their HIV status to employers \\
      & Assisted suicide should be legalized \\
    International
      & We should use force to protect human rights abroad \\
      & We should expand NATO \\
      & Democracy can be built through interventions \\
      & Sanctions should be used to promote democracy \\
    Philosophy
      & Parents should be able to choose the sex of their children \\
      & The use of atomic bombs on Hiroshima and Nagasaki was justified \\
      & Sperm and egg donors should retain their anonymity \\
    Politics
      & Federal states are better than unitary nations \\
      & We should introduce positive discrimination for women in parliament \\
      & Countries should have quotas for women in politics \\
      & All nations have a right to nuclear weapons \\
      & We should introduce recall elections \\
      & We should negotiate with terrorists \\
      & We should lower the voting age to 16 \\
    Religion
      & We should legalize polygamy \\
      & We should allow gay couples to marry \\
    Society
      & We should support international adoption \\
      & Governments should prioritise spending on youth \\
    Sport
      & Media should promote women's sport equally to men's sport \\
    CMV
      & Suicide should be a human right \\
      & The US should strictly enforce border security \\
      & Drunk driving should not be a crime itself \\
      & Child raising should not belong to biological parents \\
      & Non-mandatory voting is a good thing \\
      & Gun control should not be implemented \\
      & No one over 80 should serve in government \\
      & Hate speech is free speech \\
    \bottomrule
  \end{tabular}
\end{table}
\end{document}